\documentclass[10pt,twocolumn,letterpaper]{article}

\usepackage[pagenumbers]{cvpr} %

\usepackage{graphicx}
\usepackage{amsmath}
\usepackage{amssymb}
\usepackage{booktabs}

\usepackage{subcaption,siunitx,booktabs}
\usepackage{xcolor}
\usepackage{amsmath}
\usepackage{amssymb}
\usepackage{enumitem}

\usepackage{array}
\usepackage{caption}

\newcolumntype{L}[1]{>{\raggedright\let\newline\\\arraybackslash\hspace{0pt}}m{#1}}
\newcolumntype{C}[1]{>{\centering\let\newline\\\arraybackslash\hspace{0pt}}m{#1}}
\newcolumntype{R}[1]{>{\raggedleft\let\newline\\\arraybackslash\hspace{0pt}}m{#1}}
\usepackage{multirow}
\usepackage{graphicx}

\newcommand{\tabref}[1]{Tab.~\ref{#1}}

\newcommand{\equref}[1]{Eqn.~(\ref{#1})}

\newcommand{\colwidthA}{1.0cm}

\usepackage[pagebackref,breaklinks,colorlinks]{hyperref}

\usepackage[capitalize]{cleveref}
\crefname{section}{Sec.}{Secs.}
\Crefname{section}{Section}{Sections}
\Crefname{table}{Table}{Tables}
\crefname{table}{Tab.}{Tabs.}

\def\cvprPaperID{7867} %
\def\confName{CVPR}
\def\confYear{2022}

\begin{document}

\title{Lite Vision Transformer with Enhanced Self-Attention}

\author{Chenglin Yang\textsuperscript{1}\footnotemark, Yilin Wang\textsuperscript{2}, Jianming Zhang\textsuperscript{2}, He Zhang\textsuperscript{2}, Zijun Wei\textsuperscript{2}, Zhe Lin\textsuperscript{2}, Alan Yuille\textsuperscript{1}\\
\textsuperscript{1}Johns Hopkins University\quad\textsuperscript{2}Adobe Inc.\\
{\tt\small \{chenglin.yangw,alan.l.yuille\}@gmail.com \quad \{yilwang,jianmzha,hezhan,zwei,zlin\}@adobe.com}}

\maketitle

\renewcommand{\thefootnote}{\fnsymbol{footnote}}
\setcounter{footnote}{1} 
\footnotetext{Work done while an intern at Adobe.}
\setcounter{footnote}{0} 
\renewcommand*{\thefootnote}{\arabic{footnote}}

\begin{abstract}

Despite the impressive representation capacity of vision transformer models, current light-weight vision transformer models still suffer from inconsistent and incorrect dense predictions at local regions. We suspect that the power of their self-attention mechanism is limited in shallower and thinner networks.  
We propose Lite Vision Transformer (LVT), a novel light-weight transformer network with two enhanced self-attention mechanisms to improve the model performances for mobile deployment.
For the low-level features, we introduce Convolutional Self-Attention (CSA). Unlike previous approaches of merging convolution and self-attention, 
CSA introduces local self-attention into the convolution within a kernel of size $3 \times 3$ to enrich low-level features in the first stage of LVT. 
For the high-level features, we propose Recursive Atrous Self-Attention (RASA), which utilizes the multi-scale context when calculating the similarity map and a recursive mechanism to increase the representation capability with marginal extra parameter cost.
The superiority of LVT is demonstrated on ImageNet recognition, ADE20K semantic segmentation, and COCO panoptic segmentation. The code is made publicly available\footnote{\url{https://github.com/Chenglin-Yang/LVT}}.

\end{abstract}

\renewcommand{\colwidthA}{2.0cm}
\begin{table}[!htp]
 \centering
 \small
 \setlength{\tabcolsep}{0.0pt}
 \begin{tabular}{C{\colwidthA}C{\colwidthA}C{\colwidthA}C{\colwidthA}}
    \multicolumn{4}{c}{\includegraphics[width=0.47\textwidth]{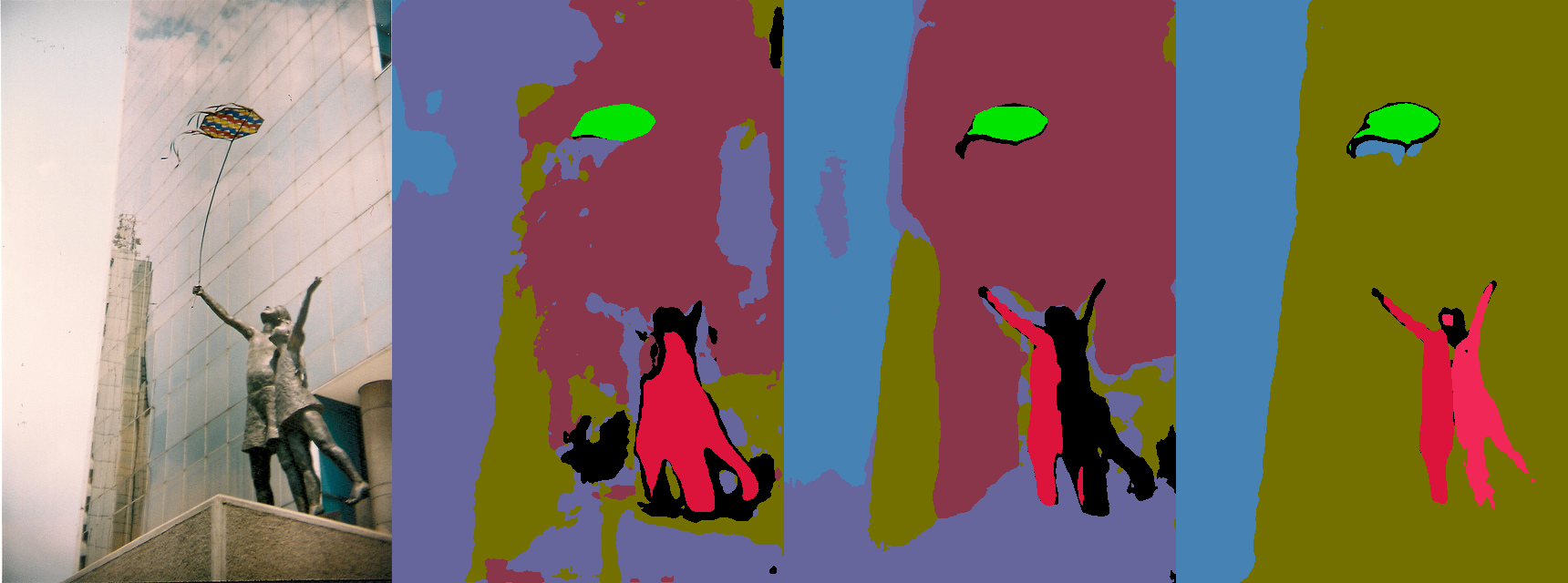}}\\
    \multicolumn{4}{c}{\includegraphics[width=0.47\textwidth]{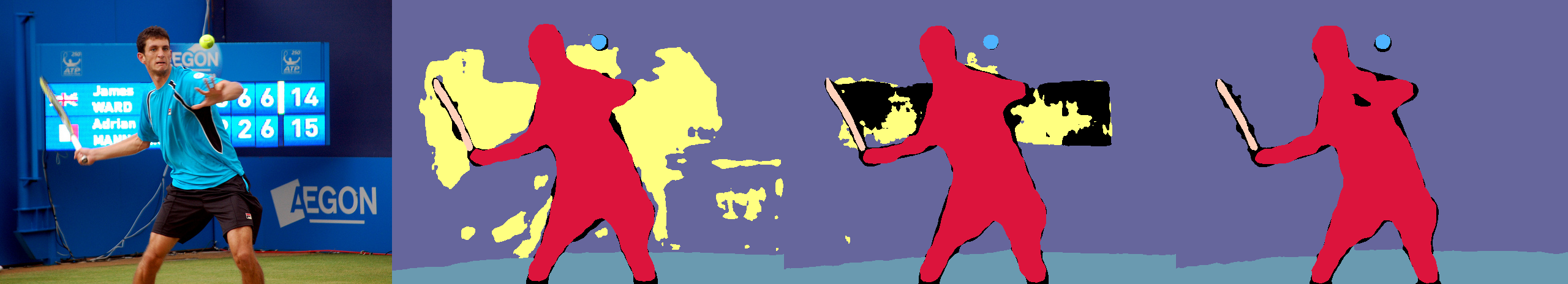}}\\
    \multicolumn{4}{c}{\includegraphics[width=0.47\textwidth]{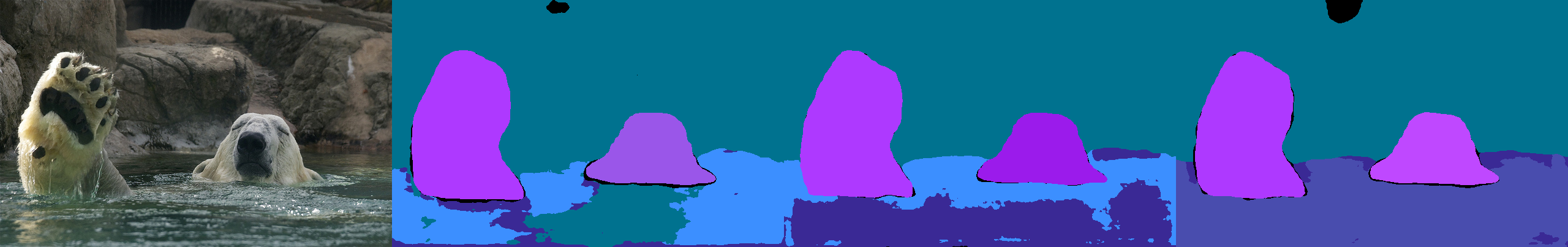}}\\
    Image & MobileNetV2 & PVTv2-B0 & LVT \\
 \end{tabular}
 \captionof{figure}{Mobile COCO panoptic segmentation. The model needs to recognize, localize, and segment both objects and stuffs at the same time. All the methods have less than $5.5$M parameters with same training and testing process under panoptic FPN framework. The only difference is the choice of encoder architecture. Lite Vision Transformer (LVT)  leads to the best results with significant improvement over the accuracy and coherency of the labels.}
 \label{fig: visual comparisons_8}
\end{table}

\section{Introduction}

Transformer-based architectures have achieved remarkable success most recently, they demonstrated superior performances on a variety of vision tasks, including visual recognition~\cite{yuan2021volo}, object detection~\cite{liu2021swin,wang2021pvtv2}, semantic segmentation~\cite{xie2021segformer,cheng2021per} and etc~\cite{wang2021max,li2021panoptic,wan2021high}.

Inspired by the success of the self-attention module in the Natural Language Processing (NLP) community \cite{vaswani2017attention}, Dosovitskiy~\cite{dosovitskiy2020image} first propose a transformer-based network for computer vision, where the key idea is to split the image into patches so that it can be linearly embedded with positional embedding. To reduce the computational complexity introduced by  \cite{dosovitskiy2020image},  Swin-Transformer~\cite{liu2021swin} upgrades the architecture by limiting the computational cost of self-attention with local non-overlapping windows. Additionally, the hierarchical feature representations are introduced to leverage features from different scales for better representation capability. On the other hand, PVT~\cite{wang2021pyramid,wang2021pvtv2} proposes spatial-reduction attention (SRA) to reduce the computational cost. It also removes the positional embedding by inserting depth-wise convolution into the feed forward network (FFN) which follows the self-attention layer in the basic transformer block. Both Swin-Transformer and PVT have demonstrated their effectiveness for downstream vision tasks. However, when scaling down the model to a mobile friendly size, there is also a significant performance degradation.

In this work, we focus on designing a light yet effective vision transformer for mobile applications~\cite{sandler2018mobilenetv2}. More specifically, we introduce a Lite Vision Transformer (LVT) backbone with two novel self-attention layers to pursue both the performance and compactness. LVT follows a standard four-stage structure~\cite{he2016deep,liu2021swin,wang2021pyramid} but has similar parameter size to existing mobile networks such as MobileNetV2~\cite{sandler2018mobilenetv2} and PVTv2-B0~\cite{wang2021pvtv2}. 

Our first improvement of self-attention is named Convolutional Self-Attention (CSA).
The self-attention layers~\cite{bahdanau2014neural,wang2018non,ramachandran2019stand,hu2019local,zhao2020exploring} are the essential components in vision transformer, as self-attention captures both short- and long-range visual dependencies. However, identifying the locality is another significant key to success in vision tasks. For example, the convolution layer is a better layer to process low-level features~\cite{dai2021coatnet}.  Prior arts have been proposed to combine convolution and self-attention with the global receptive field~\cite{dai2021coatnet,wu2021cvt}. 
Instead, we introduce the local self-attention into the convolution within the kernel of size $3 \times 3$. CSA is proposed and used in the first stage of LVT.
As a result of CSA, LVT has better generalization ability as it enriches the low-level features over existing transformer models. As shown in Fig ~\ref{fig: visual comparisons_8}, compared to PVTv2-B0 ~\cite{wang2021pvtv2}, LVT is able to generate more coherent labels in local regions.

On the other hand, the performances of lite models are still limited by the parameter number and model depth~\cite{xie2021segformer}.  We further propose to increase the representation capacity of lite transformers by Recursive Atrous Self-Attention (RASA) layers.  As shown in Fig ~\ref{fig: visual comparisons_8}, LVT results have better semantic correctness due to such effective representations. Specifically, RASA incorporates two components with weight sharing mechanisms. The first one is Atrous Self-Attention (ASA). It utilizes the multi-scale context with a single kernel when calculating the similarities between the query and key. The second one is the recursion pipeline. Following standard recursive network~\cite{elman1990finding,jordan1997serial}, we formalize RASA as a recursive module with ASA as the activation function. It increases the network depth without introducing additional parameters.

Experiments are performed on ImageNet~\cite{russakovsky2015imagenet} classification, ADE20K~\cite{zhou2017scene} semantic segmentation and COCO~\cite{lin2014microsoft} panoptic segmentation to evaluate the performance of LVT as a generalized vision model backbone. Our main contributions are summarized in the following: 

\begin{itemize}[noitemsep]
    \item We propose Convolutional Self-Attention (CSA). Unlike previous methods of merging global self-attention with convolution, CSA integrates local self-attention into the convolution kernel of size $3 \times 3$. It is proposed to process low-level features by including both dynamic kernels and learnable filters.
    \item We propose Recursive Atrous Self-Attention (RASA). It comprises two parts. The first part is Atrous Self-Attention (ASA) that captures the multi-scale context in the calculation of similarity map in self-attention. The other part is the recursive formulation with ASA as the activation function. RASA is proposed to increase the representation ability with marginal extra parameter cost.
    \item We propose Lite Vision Transformer (LVT) as a light-weight transformer backbone for vision models. LVT contains four stages and adopts CSA and RASA in the first and last three stages, respectively. The superior performances of LVT are demonstrated in ImageNet recognition, ADE20K semantic segmentation, and COCO panoptic segmentation.
\end{itemize}

\begin{table*}[!htp]
 \centering
 \small
 \setlength{\tabcolsep}{0.0pt}
 \begin{tabular}{c}
    \includegraphics[width=0.9\textwidth]{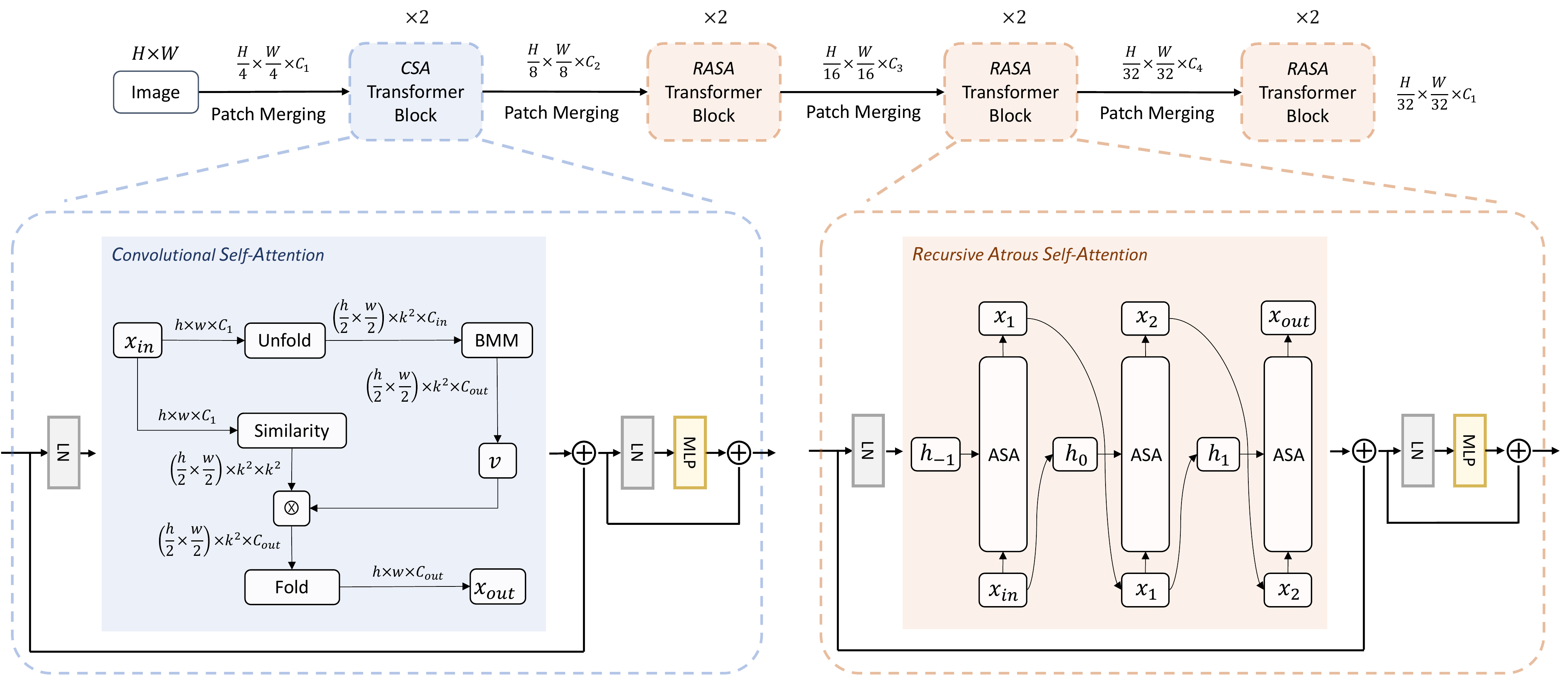}
 \end{tabular}
 \captionof{figure}{Lite Vision Transformer (LVT). The top row represents the overall structure of LVT. The bottom left and right parts visualize the proposed Convolutional Self-Attention (CSA) and Recursive Atrous Self-Attention (RASA). $H,W$ represents the height and width of the image. $C$ is the feature channel. The output resolution of each module is shown. Both the \texttt{Unfold} and \texttt{Fold} operations has a stride of $2$. BMM stands for batched matrix multiplication, which corresponds to $W_{i \rightarrow j} x_{j}$ in~\equref{equ: analyze_convolution} with the batch dimension being the number of spatial locations in a local window.
 ASA stands for the proposed Atrous Self-Attention.}
 \label{fig: overall}
\end{table*}

\section{Related Work}

\noindent\textbf{Vision Transformer.}
ViT~\cite{dosovitskiy2020image} is the first vision transformer that proves that the NLP transformer~\cite{vaswani2017attention} architecture can be transferred to the image recognition task with excellent performances. The image is split into a sequence of patches that is linearly embedded as the token inputs for ViT. After ViT, a series of improving methods are proposed.  
For training, DeiT~\cite{touvron2021training} introduces the knowledge distillation strategy for transformer. For the tokenization, T2T-ViT~\cite{yuan2021tokens} proposes T2T module to recursively aggregate neighboring tokens into one token to enrich local structure modeling. TNT~\cite{han2021transformer} further splits the tokens into smaller tokens, extract features from them to be integrated with the normal token features. For position embedding, CVPT~\cite{chu2021conditional} proposes dynamic position encoding that generalizes to images with arbitrary resolutions. For the multi-scale processing, Twins~\cite{chu2021twins} investigates the combination of local and global self-attention. CoaT~\cite{xu2021co} introduces convolution into the position embedding and investigates the cross attention among the features at various scales from different stages. Cross ViT~\cite{chen2021crossvit} proposes dual-path transformers that process tokens of different scales and adopts a token fusion module based on cross attention. For hierarchical design, Swin-Transformer~\cite{liu2021swin} and PVT~\cite{wang2021pyramid} both adopt four-stage design and gradually downsamples the feature maps which is beneficial to the downstream vision tasks.

\noindent\textbf{Combining Convolution and Self-Attention.}
There are four categories of methods. 
The first one is incorporating the position embedding in self-attention with convolution, including CVPT~\cite{chu2021conditional} and CoaT~\cite{xu2021co}. The second one is applying convolution before self-attention, including CvT~\cite{wu2021cvt},  CoAtNet~\cite{dai2021coatnet} and BoTNet~\cite{srinivas2021bottleneck}. The third one is inserting convolution after self-attention, including LocalViT~\cite{li2021localvit} and PVTv2~\cite{wang2021pvtv2}. The fourth one is to parallel self-attention and convolution, including AA~\cite{bello2019attention} and LESA~\cite{yang2021locally}. Different from all the above methods that merge local convolutions with global self-attention, we propose Convolutional Self-Attention (CSA) that combines self-attention and convolution both with $3 \times 3$ kernels as a powerful layer in the first stage of the model.

\noindent\textbf{Recursive Convolutional Neural Networks.}
Recursive methods have been exploited on the convolutional neural networks (CNNs) for various vision tasks. It includes image recognition~\cite{liang2015recurrent}, super resolution~\cite{kim2016deeply}, object detection~\cite{singh2018analysis,liang2015recurrent,liu2020cbnet,qiao2021detectors}, semantic segmentation~\cite{lin2017refinenet,cheng2020cascadepsp}. Unlike these methods, we investigate a recursive method in the light-weight vision transformer as a general model backbone. Specifically, we propose a recursive self-attention layer with the multi-scale query information, which improves the performance of the mobile model effectively.

\section{Lite Vision Transformer}
We propose Lite Vision Transformer (LVT), which is shown in Fig ~\ref{fig: overall}. As a backbone network for multiple vision tasks, we follow the standard four-stage design~\cite{he2016deep,liu2021swin,wang2021pyramid}. Each stage performs one downsampling operation and consists of a series of building blocks. Their output resolutions are from stride-$4$ to stride-$32$ gradually. Unlike previous vision transformers~\cite{dosovitskiy2020image,liu2021swin,wang2021pyramid,yuan2021volo}, LVT is proposed with limited amount of parameters and two novel self-attention layers. 
The first one is the Convolutional Self-Attention layer which has a $3 \times 3$ sliding kernel and is adopted in the first stage. The second one is the Recursive Atrous Self-Attention layer which has a global kernel and is adopted in the last three stages.

\subsection{Convolutional Self-Attention (CSA)}

\begin{table}[t]
 \centering
 \small
 \setlength{\tabcolsep}{0.0pt}
 \begin{tabular}{c}
    \includegraphics[width=0.45\textwidth]{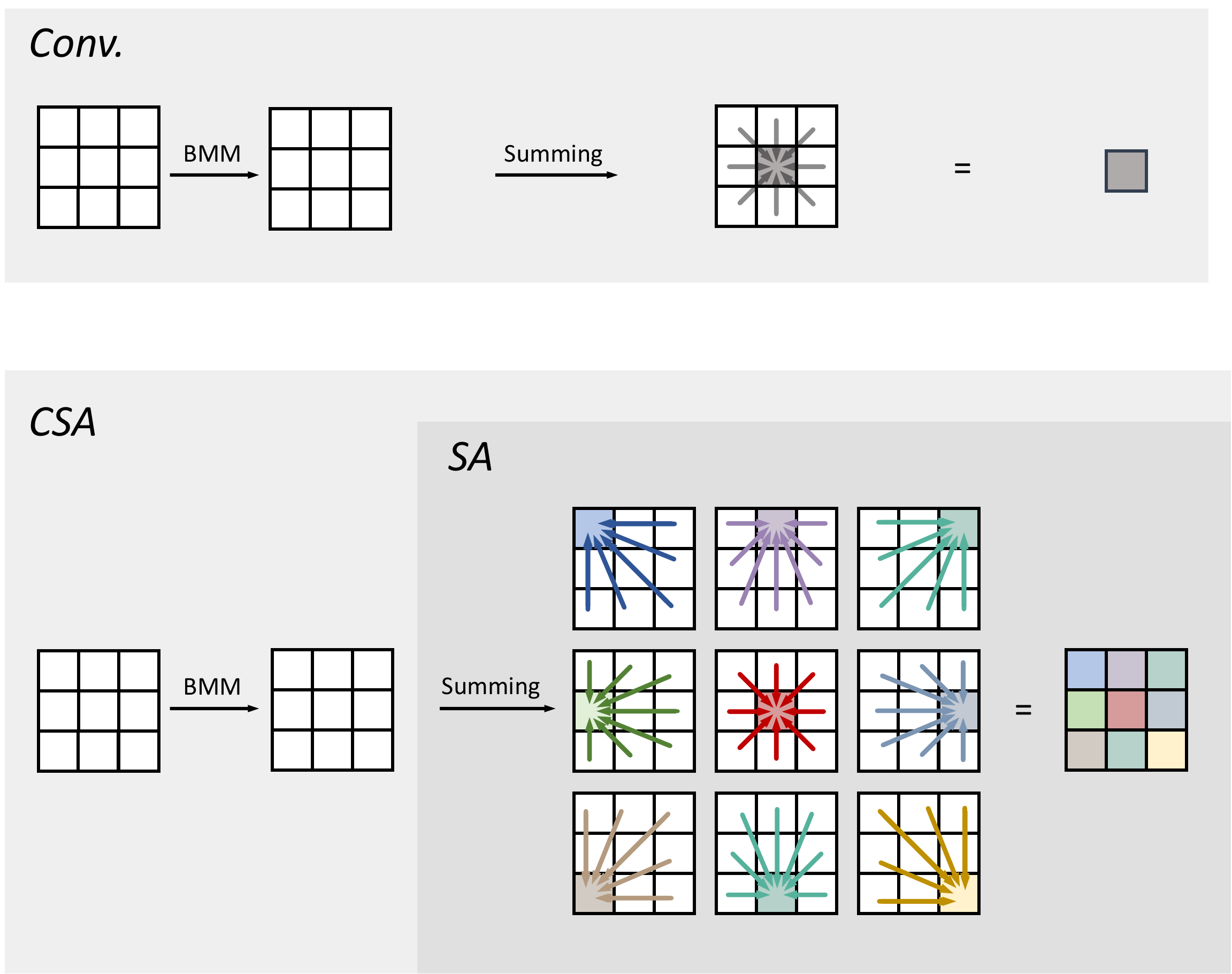}\\
 \end{tabular}
 \captionof{figure}{Illustration of Convolutional Self-Attention (CSA) in a $3 \times 3$ local window. The outputs of convolution and CSA are $1 \times 1$ and $3 \times 3$, respectively.
 Mathematically, convolution comprises two procedures: the batched matrix multiplication (BMM) and summation. BMM corresponds to $W_{i \rightarrow j} x_{j}$ in~\equref{equ: analyze_convolution} with the batch dimension being the number of spatial locations.
 CSA has the BMM operation, but has the same summation process as SA. It performs $9$ different input-dependent summations with the weights $\alpha$ in \equref{equ: analyze_self-attention}, which process is shown by the colored arrows and patches.
 Through this design, CSA contains both the learnable filter and dynamic kernel.}
 \label{fig: illustration_csa}
 \vspace{-1em}
\end{table}

The global receptive field benefits the self-attention layer in feature extraction. However, convolution is preferred in the early stages of vision models~\cite{dai2021coatnet} as the locality is more important in processing low-level features. Unlike previous methods of combining convolution and large kernel (global) self-attention~\cite{dai2021coatnet,wu2021cvt}, we focus on designing a window-based self-attention layer that has a $3 \times 3$ kernel and incorporates the representation of convolution.

\noindent{\bf Analyzing Convolution.}
Let $x, y \in \mathbb{R}^{d}$ be the input and output feature vectors where $d$ represents the channel number. Let $i,j \in \mathbb{R}$ index the spatial locations. Convolution is computed by sliding windows. In each window, we can write the formula of convolution as:
\begin{equation}
    \begin{split}
    y_{i} = \sum_{j \in N(i)}{W_{i \rightarrow j} x_{j}} \\
    \end{split}
    \label{equ: analyze_convolution}
\end{equation}
where $N(i)$ represents the spatial locations in this local neighborhood that is defined by the kernel centered at location $i$. $|N(i)| = k \times k$ where $k$ is the kernel size. $i \rightarrow j$ represents the relative spatial relationship from $i$ to $j$. $W_{i \rightarrow j} \in \mathbb{R}^{d \times d}$ is the projection matrix. In total, there are $|N(i)|$ $W$s in a kernel.
A $3 \times 3$ kernel consists of $9$ such matrices $W$s.

\noindent{\bf Analyzing Self-Attention.}
Self-Attention needs three projection matrices $W_{q}, W_{k}, W_{v} \in \mathbb{R}^{d \times d}$ to compute query, key and value. In this paper, we consider sliding window based self-attention. In each window, we can write the formula of self-attention as
\begin{equation}
    \begin{split}
    y_{i} &= \sum_{j \in N(i)}{\alpha_{i \rightarrow j} W_{v} x_{j}} \\
    \alpha_{i \rightarrow j} &= \frac{e^{(W_{q}x_{i})^{T}W_{k}x_{j}}}{\sum_{z \in N(i)}{e^{(W_{q}x_{i})^{T}W_{k}x_{z}}}}
    \end{split}
    \label{equ: analyze_self-attention}
\end{equation}
where $\alpha_{i \rightarrow j} \in (0, 1)$ is a scalar that controls the contribution of the value in each spatial location in the summation. $\alpha$ is normalized by softmax operation such that $\sum_{j}{\alpha_{i \rightarrow j}} = 1$. Compared with convolution with the same kernel size $k$, the number of learnable matrices is three rather than $k^{2}$. Recently, Outlook Attention~\cite{yuan2021volo} is proposed to predict $\alpha$ instead of calculating it by the dot product of query and key, and shows superior performances when the kernel size is small. We employ this calculation, and it can be written as:
\begin{equation}
    \begin{split}
    \alpha_{i \rightarrow j} &= \frac{W_{qk}x_{i}[j]}{\sum_{z \in N(i)}{W_{qk}x_{i}[z]}}
    \end{split}
    \label{equ: pred_qk_alpha}
\end{equation}
where $W_{qk} \in \mathbb{R}^{d \times k^{2}}$ and $[j]$ means $j$th element of the vector.  

\begin{table}[t]
  \small
  \centering
  \begin{tabular}{cc|c}
    \toprule
    Operations & Input dependent & Learnable filter \\
    \midrule
    Conv. &  & $\surd$ \\
    Self-Attention & $\surd$ & \\
    Conv. Self-Attention & $\surd$ & $\surd$ \\
    \bottomrule
  \end{tabular}
  \caption{Convolutional Self-Attention: Generalization of convolution and self-attention.}
  \label{tab: contolutional_self-attention}
\end{table}

\noindent{\bf Convolutional Self-Attention (CSA).}
We generalize self-attention and convolution into a unified convolutional self-attention as shown in Fig ~\ref{fig: illustration_csa}.
Its formulation is shown in the following:
\begin{equation}
    \begin{split}
    y_{i} = \sum_{j \in N(i)}{\alpha_{i \rightarrow j} W_{i \rightarrow j}} x_{j} \\
    \end{split}
\end{equation}
Both SA and CSA have the output of size $k \times k$ for a local window.
When $\alpha_{i \rightarrow j} = 1$ where all the weights are the same, CSA is the convolution for the output center.
When $W_{i \rightarrow j} = W_{v}$ where all the projection matrices are the same, CSA is self-attention. As we employ the dynamic $\alpha$ predicted by the input, as shown in~\equref{equ: pred_qk_alpha}, Outlook Attention~\cite{yuan2021volo} is a special case of CSA. CSA has a bigger capacity than Outlook Attention. 
We summarize its property in~\tabref{tab: contolutional_self-attention}. By this generalization, CSA has both input-dependent kernel and learnable filter. 
It is designed for stronger representation capability in first stage of vision transformers.

\begin{table}[t]
 \centering
 \small
 \setlength{\tabcolsep}{0.0pt}
 \begin{tabular}{c}
    \includegraphics[width=0.45\textwidth]{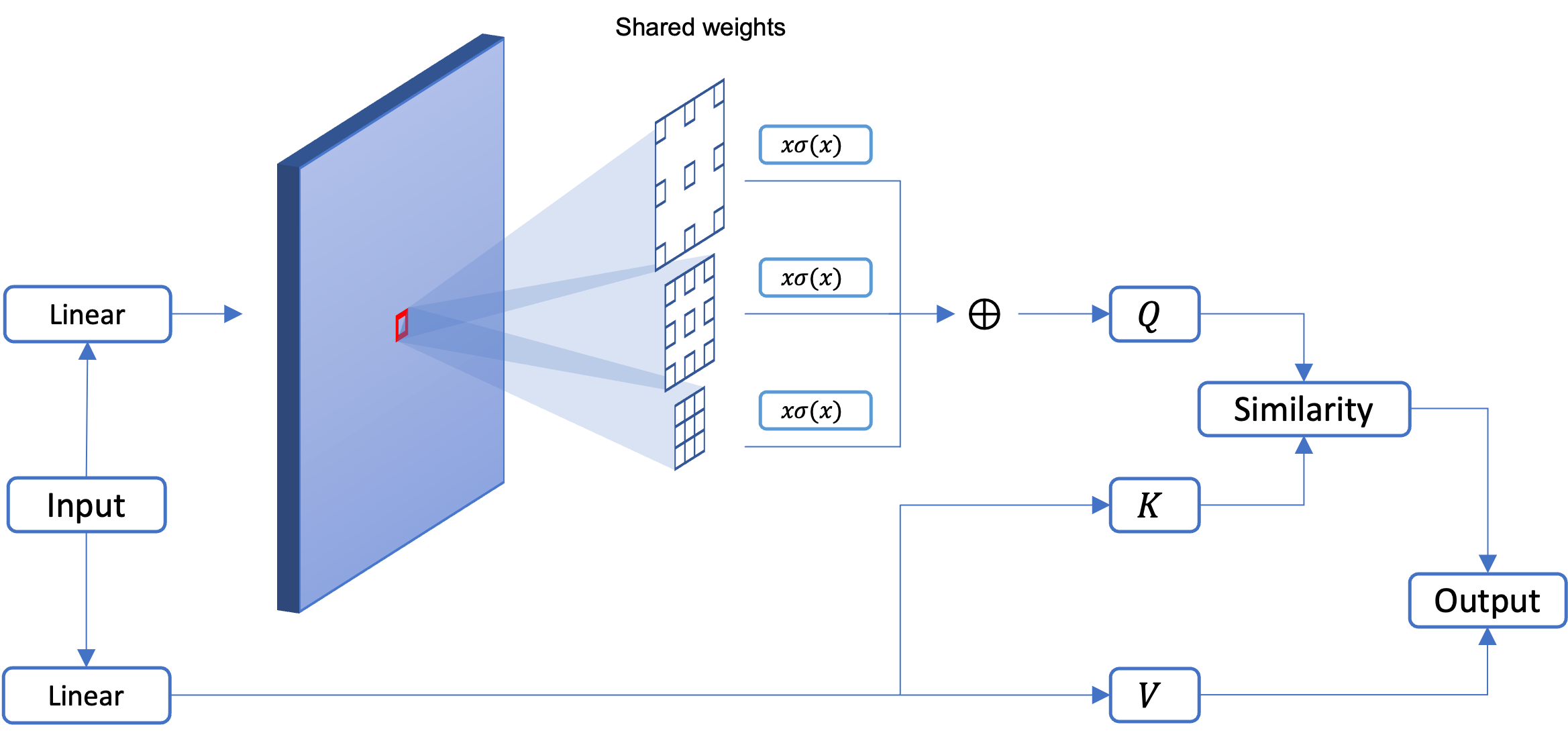}\\
 \end{tabular}
 \captionof{figure}{Illustration of Atrous Self-Attention (ASA). $Q,K,V$ stands for the query, key, and value in self-attention. ASA calculates the multi-scale query by three depth-wise convolutions after the linear projection. These convolutions share the kernel weights but have different dilation rates: $1,3,5$. Their outputs are added with the weights calculated by sigmoid function for the purpose of self-calibration. This can be implemented by the SiLU function. The multi-scale information is utilized in calculating the similarity map which weights the summation of the values.}
 \label{fig: illustration_asa}
\end{table}

\subsection{Recursive Atrous Self-Attention (RASA)}

Light-weight models are more efficient and more suitable for on-device applications~\cite{sandler2018mobilenetv2}. However, their performances are limited by the small number of parameters even with advanced model architecture~\cite{xie2021segformer}. For light-weight models, we focus on enhancing their representation capabilities with marginal increase in the number of parameters.

\noindent\textbf{Atrous Self-Attention (ASA).}
Multi-scale features are beneficial in detecting or segmenting the objects~\cite{lin2017feature,zhao2017pyramid}. 
Atrous convolution~\cite{chen2014semantic,papandreou2015modeling,chen2017rethinking} is proposed to capture the multi-scale context with the same amount of parameters as standard convolution. Weights sharing atrous convolution~\cite{qiao2021detectors} is also demonstrated in boosting model performances. 
Unlike convolution, the feature response of self-attention is a weighted sum of the projected input vectors from all spatial locations. 
These weights are determined by the similarities between the queries and keys, and represent the strength of the relationship among any pair of feature vectors. Thus we add multi-scale information when generating these weights shown in Fig ~\ref{fig: illustration_asa}. Specifically, we upgrade the calculation of the query from a $1 \times 1$ convolution to the following operation:
\begin{equation}
    Q = \sum_{r \in \{1, 3, 5\}} \text{SiLU} (\textbf{Conv}(\hat{Q}, W_{q}^{k=3}, r, g=d))
\end{equation}
where
    \begin{gather}
    \hat{Q} = \textbf{Conv}(X, W_{q}^{k=1}, r=1, g=1),\\
    \text{SiLU}(m) = m \odot \text{sigmoid}(m). 
    \end{gather}
$X,Q \in \mathbb{R}^{d \times H \times W}$ are the feature maps, and $W_{q}^{k} \in \mathbb{R}^{k^{2} \times d \times d / g}$ is the kernel weight. $H,W$ are the spatial dimensions. $d$ is the feature channels. $k$, $r$ and $g$ represent the kernel size, dilation rate, and group number of the convolution. We first use the $1 \times 1$ convolution to apply linear projection. Then we apply three convolutions that have different dilation rates but a shared kernel to capture the multi-scale contexts. The parameter cost is further reduced by setting the group number equal to the feature channel number. The parallel features of different scales are then weighted summed. We employ a self-calibration mechanism that determines the weights for each scale by their activation strength. This can be implemented by the SiLU~\cite{hendrycks2016gaussian,ramachandran2017swish}. By this design, the similarity calculation of the query and key between any pair of spatial locations in self-attention uses the multi-scale information.

\begin{table}[t]
  \centering
  \small
  \begin{tabular}{c|cccc}
    \toprule
    Architecture & Stage$1$ & Stage$2$ & Stage$3$ & Stage$4$ \\
    \midrule
    SA Type & CSA & RASA & RASA & RASA \\
    SA Kernel & $3 \times 3$ & Global & Global & Global \\
    Layer Num. & $2$ & $2$ & $2$ & $2$ \\ 
    Feature Res. & $\frac{H}{4} \times \frac{W}{4}$ & $\frac{H}{8} \times \frac{W}{8}$ & $\frac{H}{16} \times \frac{W}{16}$ & $\frac{H}{32} \times \frac{W}{32}$ \\
    Feature Dim. & $64$ & $64$ & $160$ & $256$ \\
    Heads Num. & $2$ & $2$ & $5$ & $8$ \\
    MLP Ratio & $4$ & $8$ & $4$ & $4$ \\
    SR Ratio & -- & $4$ & $2$ & $1$ \\
    \bottomrule
  \end{tabular}
  \caption{Model architecture of Lite Vision Transformer (LVT). We follow the canonical four-stage design~\cite{he2016deep}. All stages consist of transformer blocks\cite{dosovitskiy2020image}. $H,W$ are the input resolutions. SR~\cite{wang2021pyramid} mechanism is adopted for the model efficiency. We propose CSA and RASA as two enhanced Self-Attention layers (SA) to process low and high level features, respectively, in the light-weight models.}
  \label{tab: model_architecture_LVT}
\end{table}

\noindent\textbf{Resursive Atrous Self-Attention (RASA).}
For light-weight models, we intend to increase their depths without increasing the parameter usage. Recursive methods have been proposed in many vision tasks for Convolutional Neural Networks (CNNs) including~\cite{liang2015recurrent,tai2017image,kim2016deeply,qiao2021detectors}. Unlike these methods, we propose a recursive method for self-attention. The design follows the pipeline of the standard recurrent networks~\cite{elman1990finding,jordan1997serial}. Together with atrous self-attention (ASA), we propose recursive atrous self-attention (RASA), and its formula can be written in the following:
\begin{equation}
    \begin{gathered}
    x_{t+1} = \text{\bf ASA}(\mathbf{F}(X_{t}, h_{t-1})) \\
    h_{t-1} = X_{t-1} \\
    X_{t} = \text{\bf ASA}(\mathbf{F}(X_{t-1}, h_{t-2})) \\
    \end{gathered}
\end{equation}
where $t$ is the step and $h \in \mathbb{R}^{d \times H \times W}$ the hidden state. We take $\text{\bf ASA}$ as the non-linear activation function. The initial hidden state $h_{-1} = \mathbf{0}$. $\mathbf{F}(X, h) = W_{F}X + U_{F}h$ is the linear function combining the input and hidden state. $W_{F},U_{F}$ are the projection weights. However, we empirically find that setting $W_{F}=1,U_{F}=1$ provides the best performances and avoids introducing extra parameters. We set the recursion depth as two in order to limit the computation cost.

\begin{table}[t]
  \centering
  \small
  \begin{tabular}{l|c|c|c}
    \toprule
    \multirow{2}{*}{Models} & \multicolumn{1}{c|}{Top-1} & \multicolumn{1}{c|}{Params}  & {FLOPs} \\
    {} & ($\%$) & (M) & (G)  \\
    \toprule
    MobileNetV2~\cite{sandler2018mobilenetv2} & $71.9$ & $3.5$ & $0.3$ \\
    PVTv2-B0~\cite{wang2021pvtv2} & $70.5$ & $3.4$/$3.7$ & $0.6$ \\
    \textbf{LVT} & $74.8$ & $3.4$/$5.5$ & $0.9$ \\
    \midrule
    ResNet50~\cite{he2016deep} & $76.1$ & $25.6$ & $4.1$ \\
    ResNeXt50-32x4d~\cite{xie2017aggregated} & $77.6$ & $25.0$ & $4.3$\\
    SENet50~\cite{hu2018squeeze} & $77.7$ & $28.1$ & $3.9$ \\
    RegNetY-4G~\cite{radosavovic2020designing} & $80.0$ & $21.0$ & $4.0$ \\
    DeiT-Small/16~\cite{touvron2021training} & $79.9$ & $22.1$ & $4.6$ \\
    CPVT-S-GAP~\cite{chu2021conditional} & $81.5$ & $23.0$ & -- \\
    T2T-ViT\textsubscript{$t$}-14~\cite{yuan2021tokens} & $81.7$ & $21.5$ & $6.1$ \\
    DeepViT-S~\cite{zhou2021deepvit} & $82.3$ & $27.0$ & $6.2$ \\
    ViP-Small/7~\cite{hou2021vision} & $81.5$ & $25.0$ & -- \\
    PVTv1-Small~\cite{wang2021pyramid} & $79.8$ & $24.5$ & $3.8$ \\
    TNT-S~\cite{han2021transformer} & $81.5$ & $23.8$ & $5.2$ \\
    CvT-13~\cite{wu2021cvt} & $81.6$ & $20.0$ & $4.5$ \\
    CoaT-Lite Small~\cite{xu2021co} & $81.9$ & $20.0$ & $4.0$ \\
    Twins-SVT-S~\cite{chu2021twins} & $81.7$ & $24.0$ & $2.9$ \\
    CrossViT-15~\cite{chen2021crossvit} & $82.3$ & $28.2$ & $6.1$ \\
    CvT-21~\cite{wu2021cvt} & $82.5$ & $32.0$ & $7.1$ \\
    BoTNet-S1-59~\cite{srinivas2021bottleneck} & $81.7$ & $33.5$ & $7.3$ \\
    RegNetY-8G~\cite{radosavovic2020designing} & $81.7$ & $39.0$ & $8.0$ \\
    T2T-ViT\textsubscript{$t$}-19~\cite{yuan2021tokens} & $82.2$  & $39.2$ & $9.8$ \\
    \hline
    Swin-T~\cite{liu2021swin} & $81.2$ & $28.0$ & $4.5$ \\
    PVTv2-B2~\cite{wang2021pvtv2} & $82.0$ & $24.8$/$25.4$ & $4.0$ \\
    \textbf{LVT\_scaled up} & $\bf 83.3$ & $24.8$/$32.2$ & $5.2$ \\
    \bottomrule
  \end{tabular}
  \caption{ImageNet Classification. Params: encoder (transferable to other tasks) / encoder + head. 
  Following MobileNetV2~\cite{sandler2018mobilenetv2} and PVTv2-B0~\cite{wang2021pvtv2}, we limit the parameter size of the encoder less than $3.5$M. In order to compare LVT with other classifiers, we scale LVT to the size of the canonical network, ResNet50~\cite{he2016deep}. We can observe that LVT shows superior performances. 
  }
  \label{tab: imagenet_classification}
\end{table}

\begin{table*}[t]
  \small
  \centering
  \begin{tabular}{cccccc}
    \toprule
    Method & Encoder & mIoU & Params (M)  & FLOPs (G) & FPS ($512$) \\
    \midrule
    FCN~\cite{long2015fully} & MobileNetV2~\cite{sandler2018mobilenetv2} & $19.7$ & $9.8$ & $39.6$ & $64.4$ \\
    PSPNet~\cite{zhao2017pyramid} & MobileNetV2~\cite{sandler2018mobilenetv2} & $ 29.6$ & $13.7$ & $52.9$ & $ 57.7$ \\
    DeepLabV3+~\cite{chen2018encoder} & MobileNetV2~\cite{sandler2018mobilenetv2} & $ 34.0$ & $15.4$ & $69.4$ & $43.1$ \\
    \hline
    SegFormer~\cite{xie2021segformer} & MiT-B0~\cite{xie2021segformer} & $37.4$ & $3.8$ & $8.4$ & $ 50.5$ \\
    SegFormer~\cite{xie2021segformer} & \textbf{LVT} & $\bf 39.3$ & $3.9$ & $10.6$ & $45.5$ \\
    \bottomrule
  \end{tabular}
  \caption{Mobile ADE20K semantic segmentation. We report the results for the single-scale input. The FPS is calculated on the $2\rm{,}000$ images whose short sides are rescaled to $512$ with the aspect ratio unchanged. It is observed that LVT achieves significant performance gain compared with previous SOTA mobile semantic segmentation models.}
  \label{tab: ade20k_semantic_segmentation}
\end{table*}

\begin{table*}[t]
  \centering
  \small
  \begin{tabular}{cc|ccc|ccc|ccc}
    \toprule
    \multirow{2}{*}{Method} & \multirow{2}{*}{Backbone} & \multicolumn{3}{c|}{COCO \texttt{val}} & \multicolumn{3}{c|}{COCO \texttt{test-dev}} & Params & FLOPs & FPS \\
    {} & {} & PQ & PQ\textsuperscript{th} & PQ\textsuperscript{st} & PQ & PQ\textsuperscript{th} & PQ\textsuperscript{st} & (M)  & (G) & ($1333$, $800$) \\
    \midrule
    Panoptic FPN~\cite{kirillov2019panoptic} & MobileNetV2~\cite{sandler2018mobilenetv2} & $36.3$ & $42.9$ & $26.4$ & $36.4$ & $43.0$ & $26.5$ & $4.1$ & $32.9$ & $35.8$ \\
    Panoptic FPN~\cite{kirillov2019panoptic} & PVTv2-B0~\cite{wang2021pvtv2} & $41.3$ & $47.5$ & $31.9$ & $41.2$ & $47.7$ & $31.5$ & $5.3$ & $49.7$ & $23.5$ \\
    Panoptic FPN~\cite{kirillov2019panoptic} & \textbf{LVT} & $\bf 42.8$ & $\bf 49.5$ & $\bf 32.6$ & $\bf 43.0$ & $\bf 49.9$ & $\bf 32.6$ & $5.4$ & $56.4$ & $20.4$ \\
    \bottomrule
  \end{tabular}
  \caption{Mobile COCO panoptic segmentation. The FPS is calculated on the $2\rm{,}000$ high-resolution images. They are rescaled such that the maximum length does not exceed $1333$ and the minimum length $800$. The aspect ratio is kept. It is observed that LVT achieves significant performance improvement over previous SOTA mobile encoders for panoptic segmentation.}
  \label{tab: coco_val_panoptic_segmentation}
\end{table*}

\subsection{Model Architecture}

The architecture of LVT is shown in ~\tabref{tab: model_architecture_LVT}. We adopt the standard four-stage design~\cite{he2016deep}. Four Overlapped Patch Embedding layers~\cite{xie2021segformer} are employed. The first one downsamples the image into stride-$4$ resolution. The other three downsample the feature maps to the resolution of stride-$8$, stride-$16$, and stride-$32$. All stages comprise the transformer blocks~\cite{dosovitskiy2020image}. Each block contains the self-attention layer followed by an MLP layer. CSA is embedded in the stage-$1$ while RASA in the other stages. They are enhanced self-attention layers proposed to process local and global features in LVT.

\section{Experiments}

\subsection{ImageNet Classification}

\begin{table*}[t]
  \centering
  \small
  \begin{tabular}{c|cc|c|c||c|c|c}
    \toprule
    Tasks & \multicolumn{4}{c||}{ImageNet Classification} & \multicolumn{3}{c}{ADE20K Sementic Segmentation}  \\
    \midrule
    \multirow{2}{*}{Methods} & \multicolumn{2}{c|}{Accuracy (\%)} & \multicolumn{1}{c|}{Params}  & FLOPs & \multirow{2}{*}{mIoU} & Params & FLOPs \\
    {} & Top-1 & Top-5 & (M) & (G) &  & (M) & (G)  \\
    \midrule
    VOLO-D0~\cite{yuan2021volo} & $74.6$ & $92.5$ & $3.9$ & $1.9$ & $39.3$ & $3.24$ & $12.28$ \\
    VOLO-D0 + CSA & $75.2$ & $\bf 92.9$ & $4.0$ & $1.9$ & $40.0$ & $3.39$ & $12.38$ \\
    VOLO-D0 + CSA + RASA & $\bf 75.6$ & $\bf 92.9$ & $4.0$ & $2.2$ & $\bf 41.0$ & $3.40$ & $15.70$ \\
    \bottomrule
  \end{tabular}
  \caption{Ablation studies. VOLO is used as the base network to add CSA and RASA, because VOLO uses the self-attention with $3 \times 3$ kernel in the first stage. By this comparison, the performance gain from local self-attention to Convolutional Self-Attention (CSA) can be clearly illustrated. It is demonstarted that both CSA and RASA significantly contributes to the performance improvement.}
  \label{tab: ablation_csa_rasa}
\end{table*}

\noindent\textbf{Dataset.}
We perform image recognition experiments on ILSVRC2012~\cite{russakovsky2015imagenet}, a popular subset of the ImageNet database~\cite{deng2009imagenet}. The training and validation sets contain $1.3\mathrm{M}$ and $50\mathrm{K}$ images, respectively. There are $1\rm{,}000$ object categories in total. The classes are distributed approximately and strictly uniformly in the training and validation sets.

\noindent\textbf{Settings.}
The training setting follows previous conventions.
We use AdamW as the optimizer~\cite{loshchilov2017decoupled}. Following previous works~\cite{jiang2021all,touvron2021training}, the learning rate is scaled based on the batch size with the formula being $\text{lr} = \frac{\text{batch\_size}}{1024} \times \text{lr\_base}$. We set lr\_base as $1.6 \times 10^{-3}$. The weight decay of $5 \times 10^{-2}$ is adopted. Stochastic depth with drop path rate being $0.1$ is employed~\cite{huang2016deep}. In total, there are $300$ training epochs. Following~\cite{yuan2021volo}, we use CutOut~\cite{zhong2020random}, RandAug~\cite{cubuk2020randaugment}, and Token Labeling~\cite{jiang2021all} as the data augmentation methods. Class attention layer~\cite{touvron2021going} is used as the post stage. Both in the training and testing phase, the input resolution is $224 \times 224$.

\noindent\textbf{Results.}
The results are shown in ~\tabref{tab: imagenet_classification}. We limit the encoder size less than $3.5$M, following MobileNet~\cite{sandler2018mobilenetv2} and PVTv2-B0\cite{wang2021pvtv2}. The encoder is our design focus as it is the backbone used by other complex tasks like detection and segmentation. In order to compare LVT with other standard models, we scale LVT to the size of ResNet50~\cite{he2016deep}, a canonical b ackbone of vision models. The high performanes of LVT for image recognition is demonstrated.

\subsection{Mobile ADE20K Semantic Segmentation}

\noindent\textbf{Dataset.}
We perform semantic segmentation task on the challenging ADE20K dataset~\cite{zhou2017scene}. There are $150$ categories in total, including $35$ stuff classes and $115$ discrete objects. The training and validation sets contain $20\rm{,}210$ and $2\rm{,}000$ images, respectively. 

\noindent\textbf{Settings.}
Previous conventions are followed. We adopt the Segformer framework~\cite{xie2021segformer} and use the MLP decoder. The LVT encoder is pretrained on ImageNet-1K without extra data. The decoder is trained from scratch. We use mmsegmentation~\cite{mmseg2020} as the codebase. 
The AdamW optimizer~\cite{loshchilov2017decoupled} with the initial learning rate being $6 \times 10^{-5}$ is used. The weight decay is set as $1 \times 10^{-2}$. The poly learning rate schedule with power being $1$ is employed. There are $160$K training iterations in total and the batch size is $16$. For data augmentation, we randomly resize the image with ratio $0.5 - 2.0$ and then perform random cropping of size $512 \times 512$. Horizontal flipping with probability $0.5$ is applied. During evaluation, we perform single-scale test.

\noindent\textbf{Results.}
The results are summarized in~\tabref{tab: ade20k_semantic_segmentation}. The FLOPs is calculated with the input resolution $512 \times 512$. The FPS is calculated on $2000$ images on a single NVIDIA V100 GPU. During inference, the images are resized such that the short side is $512$. We only use single-scale testing. The model is compact. Together with the decoder, the parameters are less than $4$M. We can observe that LVT demonstrates the best performance among all previous mobile methods for semantic segmentation.

\subsection{Mobile COCO Panoptic Segmentation}

\renewcommand{\colwidthA}{2.0cm}
\begin{table}[t]
 \centering
 \small
 \setlength{\tabcolsep}{0.0pt}
 \begin{tabular}{C{\colwidthA}C{\colwidthA}C{\colwidthA}C{\colwidthA}}
    \multicolumn{4}{c}{\includegraphics[width=0.45\textwidth]{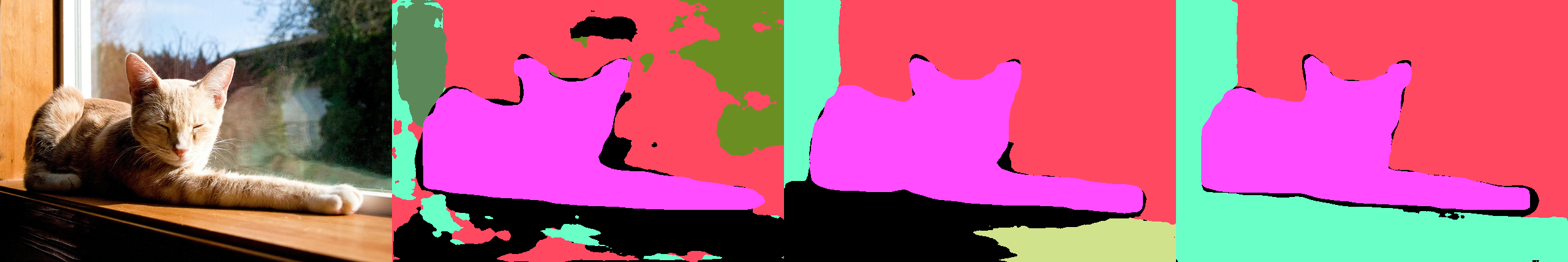}}\\
     \multicolumn{4}{c}{\includegraphics[width=0.45\textwidth]{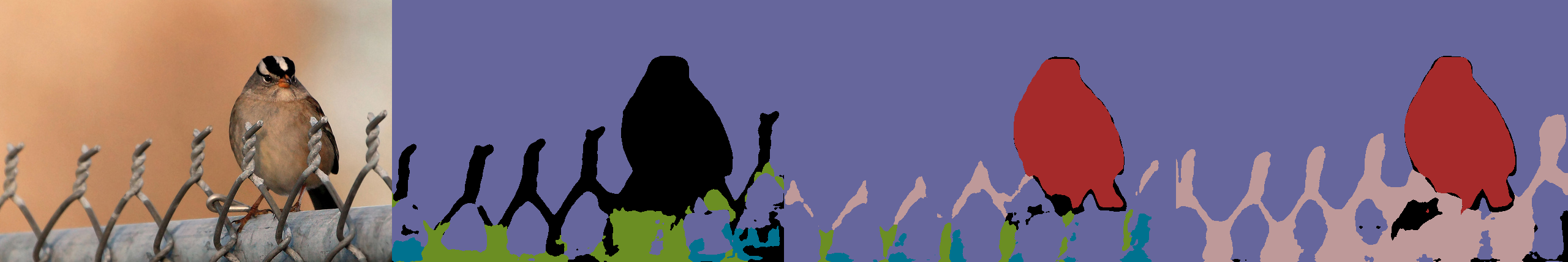}}\\
     \multicolumn{4}{c}{\includegraphics[width=0.45\textwidth]{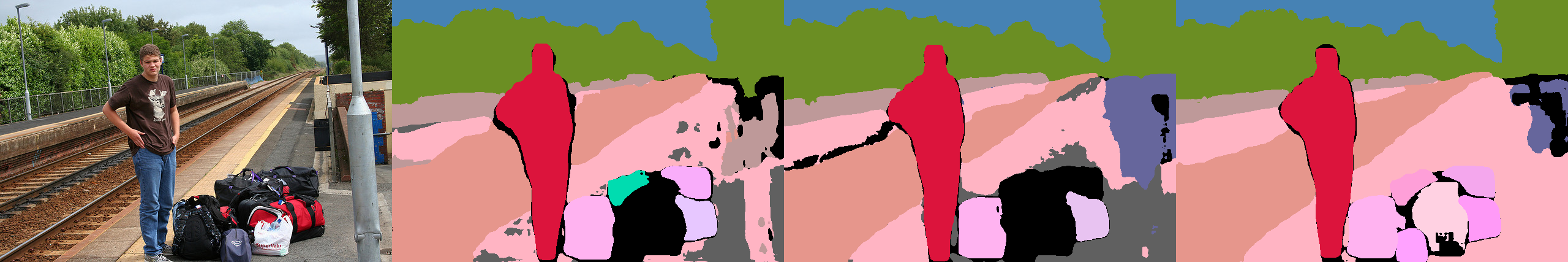}}\\
    \multicolumn{4}{c}{\includegraphics[width=0.45\textwidth]{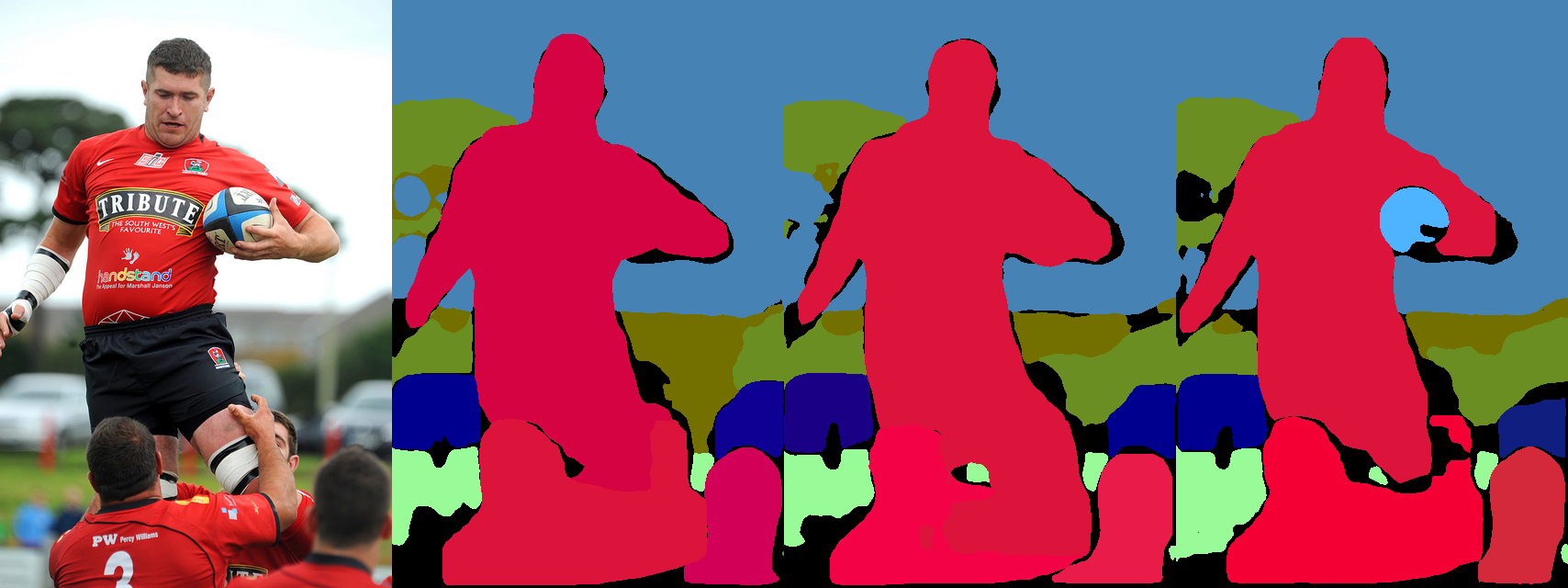}}\\
    \multicolumn{4}{c}{\includegraphics[width=0.45\textwidth]{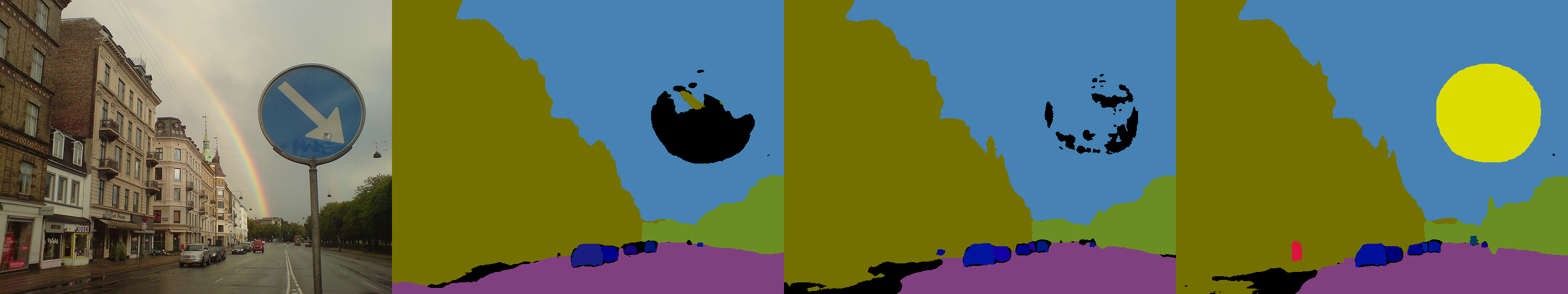}}\\
    \multicolumn{4}{c}{\includegraphics[width=0.45\textwidth]{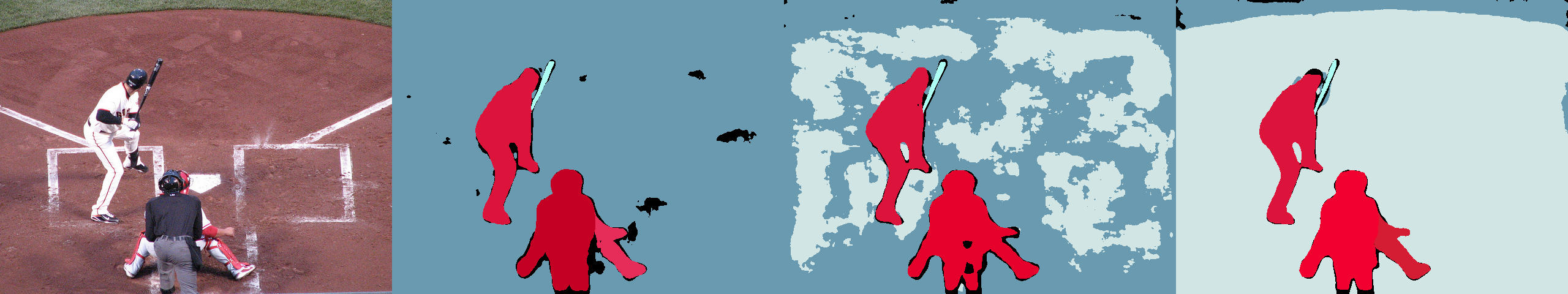}}\\
    \multicolumn{4}{c}{\includegraphics[width=0.45\textwidth]{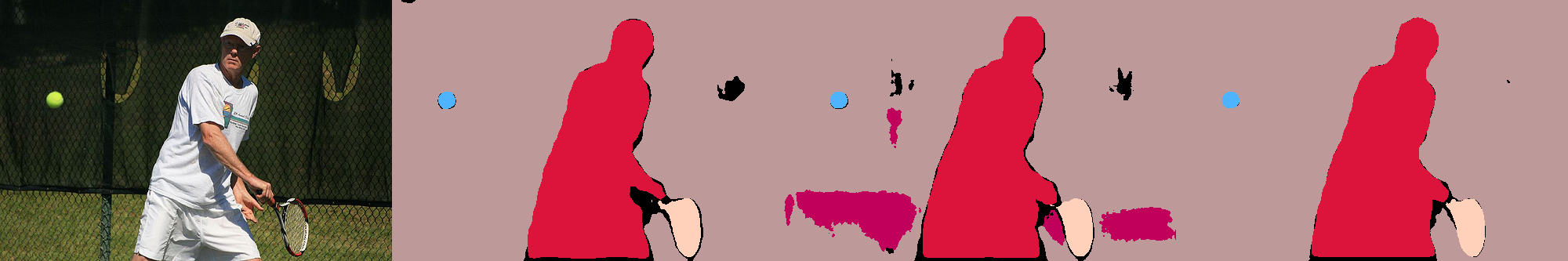}}\\
    Image & MobileNetV2 & PVTv2-B0 & LVT \\
 \end{tabular}
 \captionof{figure}{Mobile COCO panoptic segmentation. }
 \label{fig: panoptic_vis}
 \vspace{-1.5em}
\end{table}

\begin{table}[t]
  \centering
  \small
  \begin{tabular}{l|cc|c|c}
    \toprule
    \multirow{2}{*}{Models} & \multicolumn{1}{c|}{Top-1} & \multicolumn{1}{c|}{Top-5} & \multicolumn{1}{c|}{Params}  & {FLOPs} \\
    {} & ($\%$) & (M) & (G)  \\
    \toprule
    \textbf{LVT\_R1} & $73.9$ & $92.1$ & $3.4$/$5.5$ & $0.8$ \\
    \textbf{LVT\_R2} & $74.8$ & $92.6$ & $3.4$/$5.5$ & $0.9$ \\
    \textbf{LVT\_R3} & $74.6$ & $92.5$ & $3.4$/$5.5$ & $1.0$ \\
    \textbf{LVT\_R4} & $\mathbf{74.9}$ & $\mathbf{92.6}$ & $3.4$/$5.5$ & $1.1$ \\
    \bottomrule
  \end{tabular}
  \caption{Relationship of recursion times and performances on ImageNet Classification. $R$ means recursion times. The performance increases dramatically with two iterations. Considering the efficiency, we use LVT\_R2 in the main experiments.}
  \label{tab: ablation_studies_recursion}
\end{table}

\noindent\textbf{Dataset.}
We perform panoptic segmentation on COCO dataset. The 2017 split is employed. It has $118$K training images and $5$K validation images. On average, each image contains $3.5$ categories and $7.7$ instances. We choose panoptic segmentation to evaluate our methods as it unifies the object recognition, detection, localization, and segmentation at the same time. We aim to thoroughly evaluate the performance of our model.

\noindent\textbf{Settings.}
The panoptic FPN~\cite{kirillov2019panoptic} framework is adopted. All the models are trained in this framework for fair comparisons. We use mmdetection~\cite{chen2019mmdetection} as the codebase. AdamW~\cite{loshchilov2017decoupled} optimizer with the initial learning rate $3 \times 10^{-4}$ is used. The weight decay is $1 \times 10^{-4}$. The $3 \times$ schedule is employed. There are $36$ training epochs in total, the learning rate is decayed by $10$ times after $24$ and $33$ epochs. We adopt multi-scale training. During training, the images are randomly resized. The maximum length does not exceed $1333$. The maximum allowable length of the short side is randomly sampled in the range of $640 - 800$. Random horizontal flipping with probability $0.5$ is applied. During testing, we perform single-scale testing. 

\noindent\textbf{Results.}
The results are shown in~\tabref{tab: coco_val_panoptic_segmentation}. The FLOPs is calculated on the input resolution $1200 \times 800$. During the inference, all the images are resized such that the large side is not larger than $1333$ and the short side is less than $800$. The FLOPs are calculated on $2000$ high-resolution images with a single NVIDIA V100 GPU. The whole model including the decoder takes less than $5.5$M parameters. We can observe the superiority of LVT compared with the previous state of the art encoders for mobile panoptic segmentation.

\renewcommand{\colwidthA}{2.7cm}
\begin{table}[!ht]
 \centering
 \small
 \setlength{\tabcolsep}{0.0pt}
 \begin{tabular}{C{\colwidthA}C{\colwidthA}C{\colwidthA}}
    \multicolumn{3}{c}{\includegraphics[width=0.45\textwidth]{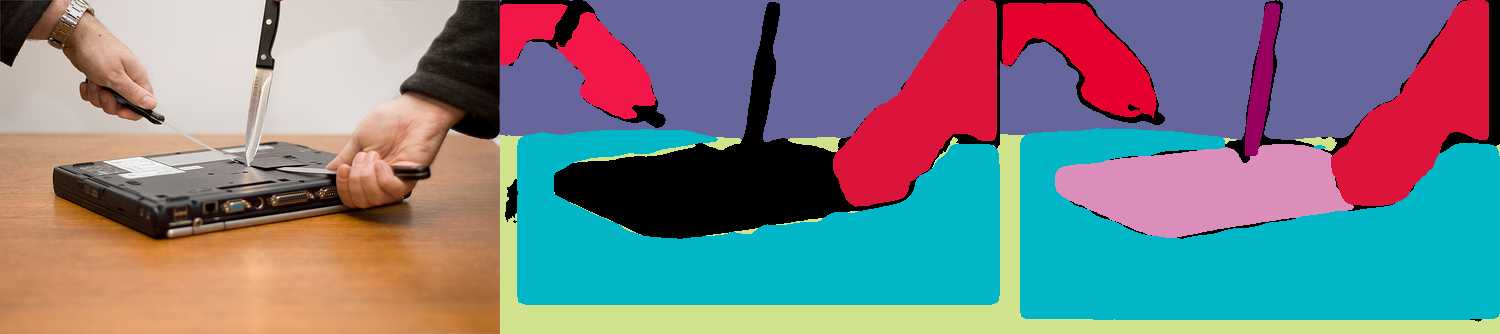}}\\
    \multicolumn{3}{c}{\includegraphics[width=0.45\textwidth]{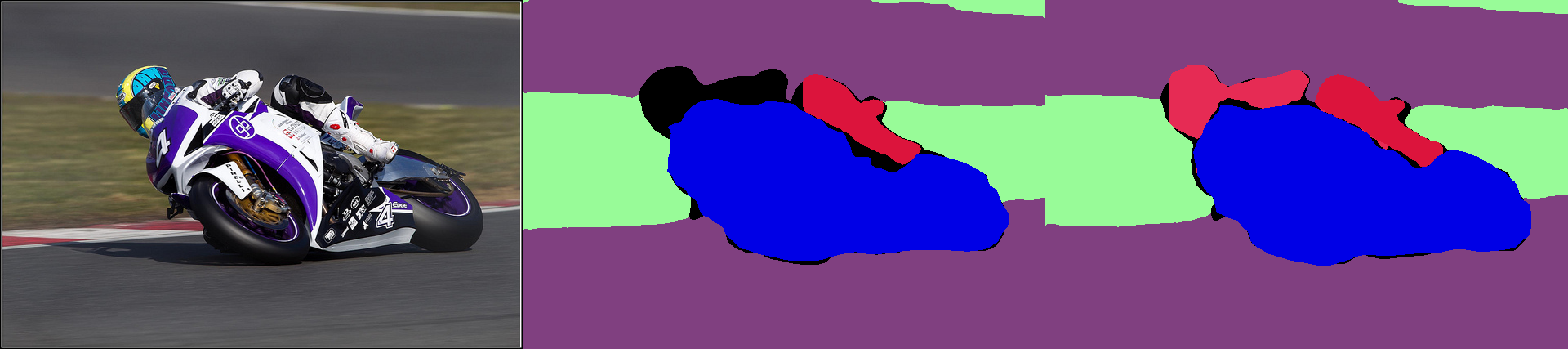}} \\
    \multicolumn{3}{c}{\includegraphics[width=0.45\textwidth]{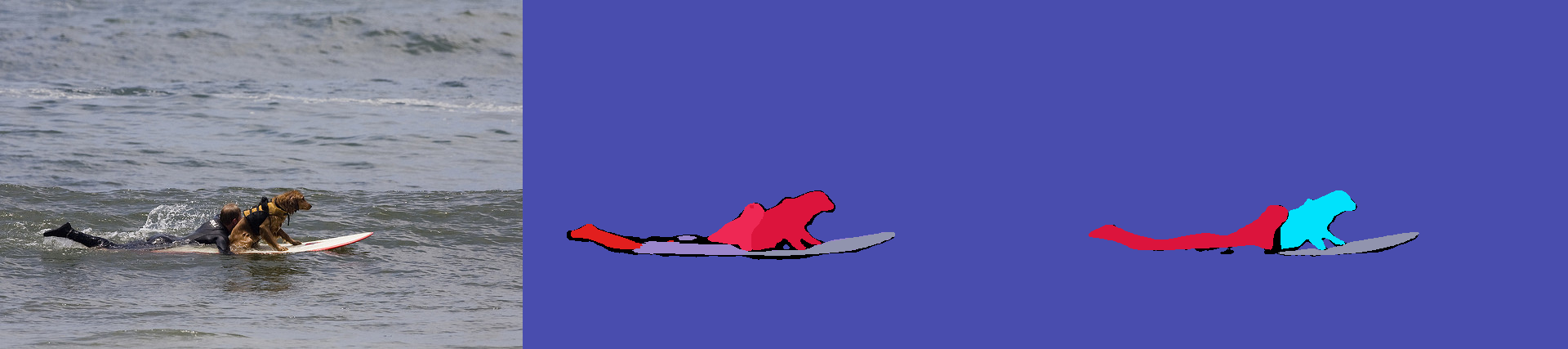}} \\
    \multicolumn{3}{c}{\includegraphics[width=0.45\textwidth]{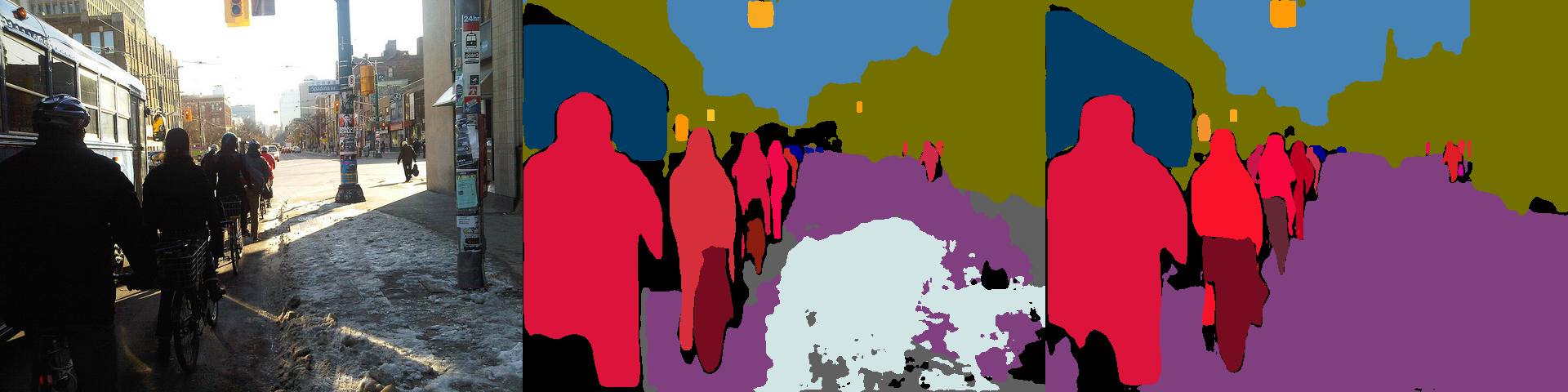}} \\
    \multicolumn{3}{c}{\includegraphics[width=0.45\textwidth]{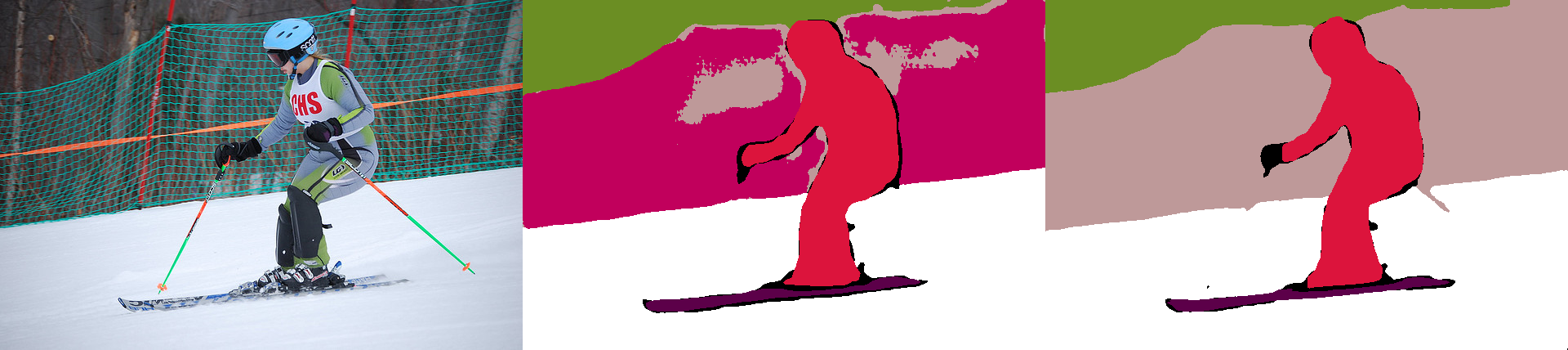}} \\
    Image & w/o CSA & w.~CSA \\
    \multicolumn{3}{c}{\includegraphics[width=0.45\textwidth]{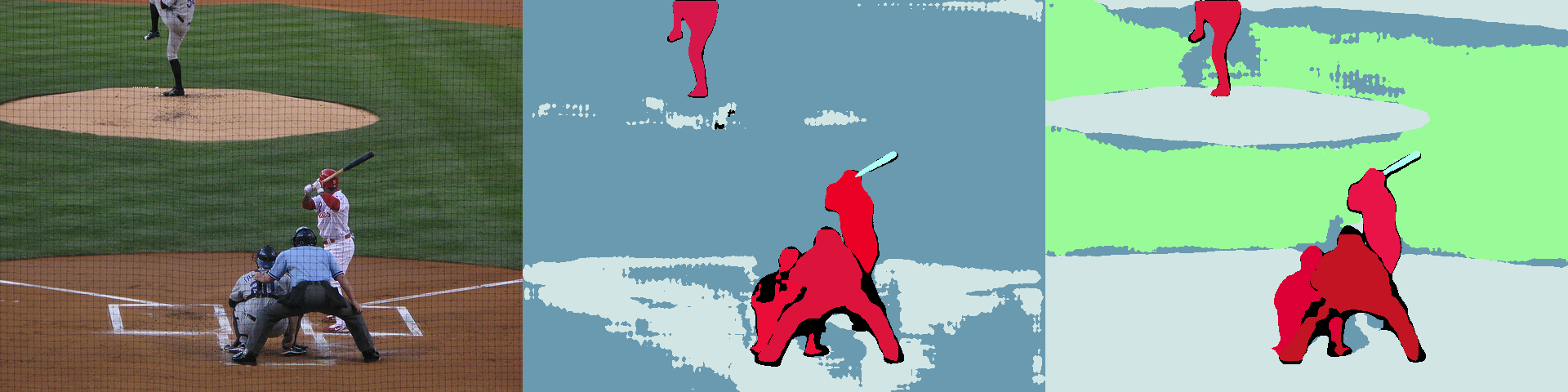}}\\
    \multicolumn{3}{c}{\includegraphics[width=0.45\textwidth]{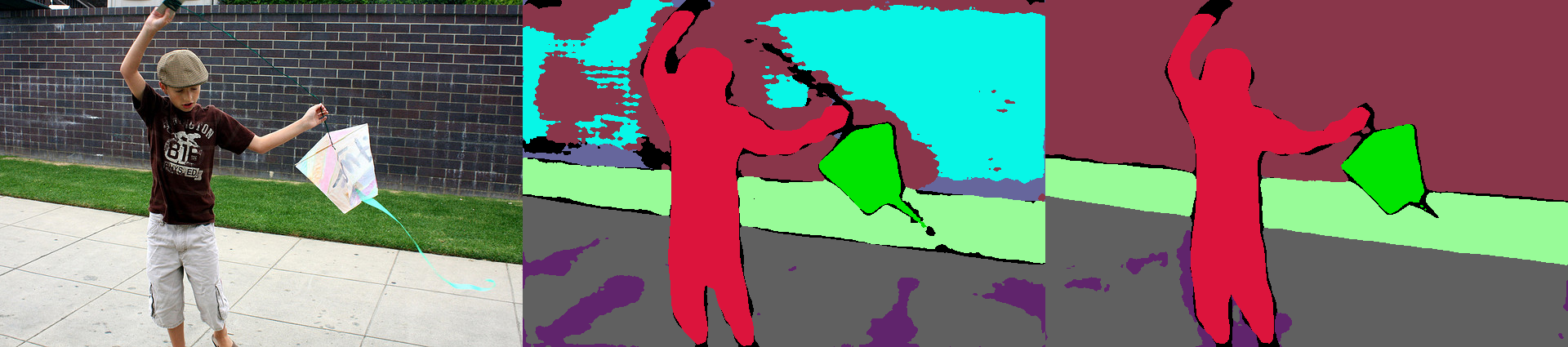}}\\
    \multicolumn{3}{c}{\includegraphics[width=0.45\textwidth]{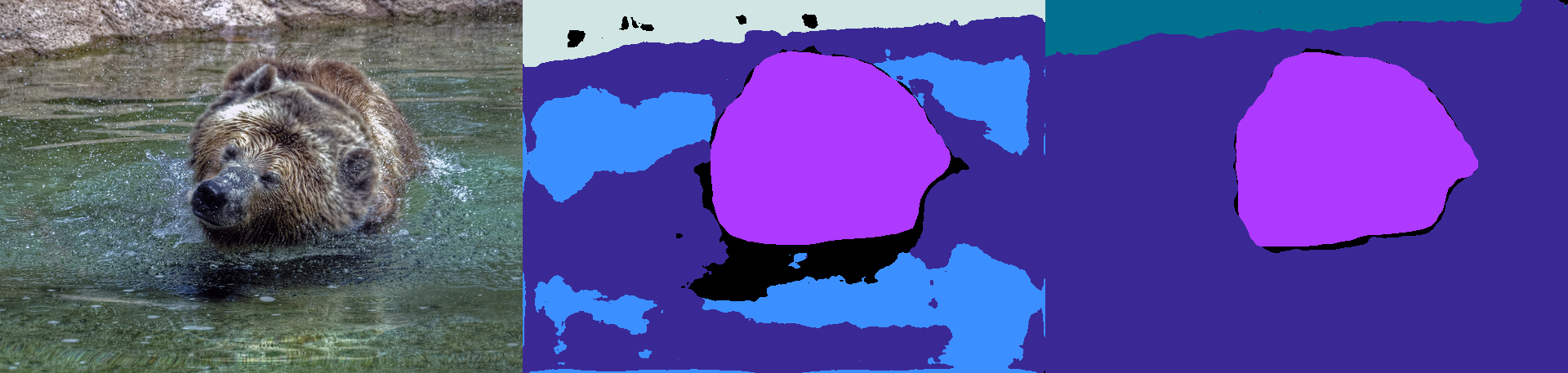}}\\
    \multicolumn{3}{c}{\includegraphics[width=0.45\textwidth]{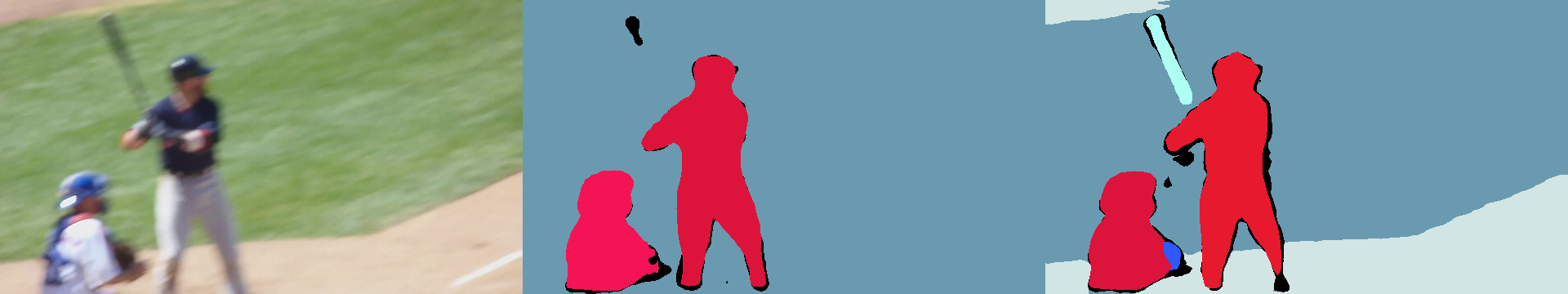}}\\
    \multicolumn{3}{c}{\includegraphics[width=0.45\textwidth,height=0.19\textwidth]{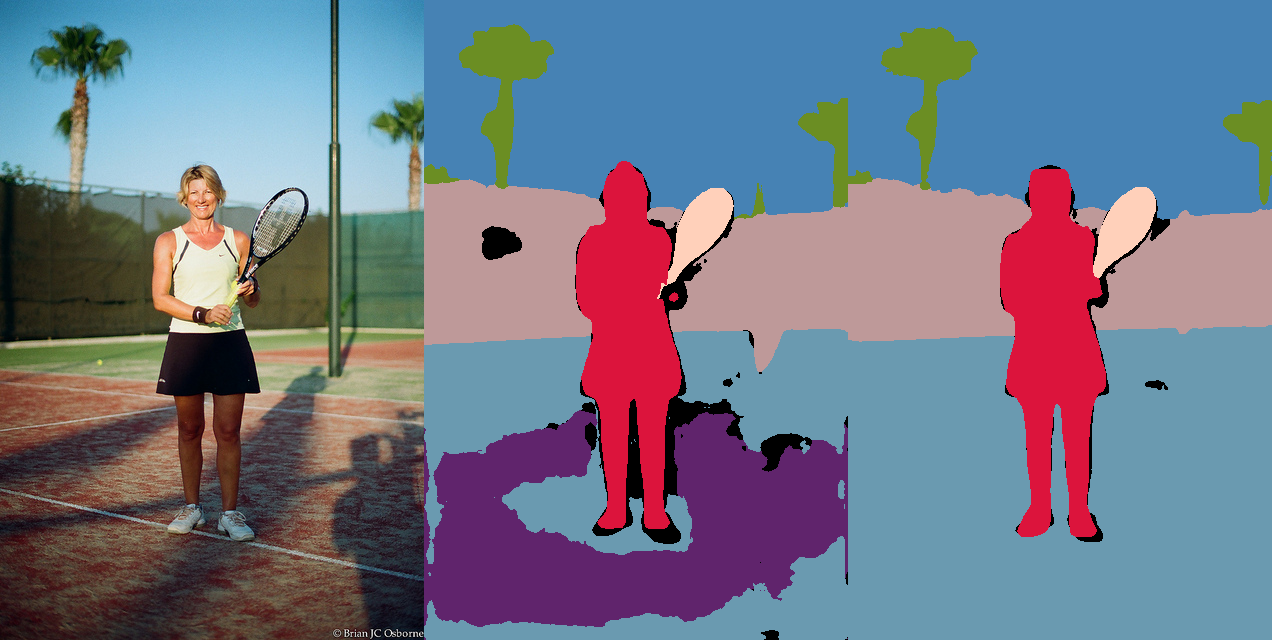}}\\
    Image & w/o RASA & w.~RASA \\
 \end{tabular}
 \captionof{figure}{Visual effectiveness of CSA and RASA in LVT.}
 \label{fig: ablation_raw-csa-rasa}
 \vspace{-1em}
\end{table}
\section{Ablation Studies}

\subsection{Recursion Times of RASA}

In this section, we investigate the relationship between the recursion times and model performances. The experiments are performed on ImageNet classification. We set the recursion time from $1$ to $4$. The results are summarized in~\tabref{tab: ablation_studies_recursion}. 

\subsection{Contributions of CSA and RASA}

\noindent\textbf{Settings.}
In this section, we study the performance contributions of Convolutional Self-Attention (CSA) and Recursive Atrous Self-Attention. To this end, we build our model via the recently proposed VOLO which employed small kernel self-attention in the first stage. As VOLO is demonstrated as a powerful backbone in image recognition and semantic segmentation, we perform experiments on ImageNet and ADE20K. In order to perform the comparisons in the mobile setting, we scale VOLO to have a parameter size $4.0$M. Specifically, we set the layer number of each stage to be $2$, and adjust the feature dimensions to be $96, 192, 192, 192$. All the other settings are kept unchanged. 

\noindent\textbf{Results.}
The results are reported in~\tabref{tab: ablation_csa_rasa}. For ImageNet classification, the input resolution in both the training and testing is $224 \times 224$. For ADE20K semantic segmentation, we insert the VOLO and LVT with the MLP decoder, following the Segformer framework~\cite{xie2021segformer}. During testing, the short side of the image is resized to $512$. It is observed both CSA and RASA significantly contribute to the performance gain.

\section{Conclusion}

In this work, we propose a powerful light-weight transformer backbone, Lite Vision Transformer (LVT). It consists of two novel self-attention layers: Convolutional Self-Attention (CSA) and Recursive Atrous Self-Attention (RASA). They are used in the first and the last three stages of LVT to process low and high level features. We demonstrate the strong performances of LVT compared with previous mobile methods in tasks of visual recognition, semantic segmentation, and panoptic segmentation.

\noindent\textbf{Limitations.}
LVT is a light-weight model. The natural limitation is the weak representation power compared with models with large number of parameters. The focus of this work is on the mobile models. Our future work includes scaling LVT to the large powerful backbones.

{\small
\bibliographystyle{ieee_fullname}
\bibliography{egbib}
}

\end{document}